\documentclass[10pt,twocolumn,letterpaper]{article}

\usepackage{cvpr}
\usepackage{times}
\usepackage{epsfig}
\usepackage{graphicx}
\usepackage{amsmath}
\usepackage{amssymb}
\usepackage{xcolor}
\usepackage{booktabs}
\usepackage{multirow}

\usepackage[pagebackref=true,breaklinks=true,letterpaper=true,colorlinks,bookmarks=false]{hyperref}

\cvprfinalcopy 

\def\cvprPaperID{931} 

\ifcvprfinal\fi

\begin{document}

\title{Face X-ray for More General Face Forgery Detection}

\author{Lingzhi Li$^{1}$\thanks{Equal contribution}  \thanks{Work done during an internship at Microsoft Research Asia} \quad Jianmin Bao$^{2*}$\thanks{Corresponding Author} \quad Ting Zhang$^{2}$ \quad Hao Yang$^{2}$ \quad Dong Chen$^{2}$ \quad Fang Wen$^{2}$  \quad Baining Guo$^{2}$ \qquad \vspace{1pt}\\
$^{1}$Peking University  \qquad $^{2}$Microsoft Research Asia\qquad\qquad\\
\hspace{0.1in}{\tt\small lilingzhi@pku.edu.cn} \qquad  {\tt\small \{jianbao,tinzhan,haya,doch,fangwen,bainguo\}@microsoft.com} \\
}

\maketitle
\thispagestyle{empty}

\begin{abstract}
In this paper we propose a novel image representation called face X-ray for detecting forgery in face images. The face X-ray of an input face image is a greyscale image that reveals whether the input image can be decomposed into the blending of two images from different sources. It does so by showing the blending boundary for a forged image and the absence of blending for a real image.  We observe that most existing face manipulation methods share a common step: blending the altered face into an existing background image. For this reason, face X-ray provides an effective way for detecting forgery generated by most existing face manipulation algorithms. Face X-ray is general in the sense that it only assumes the existence of a blending step and does not rely on any knowledge of the artifacts associated with a specific face manipulation technique. Indeed, the algorithm for computing face X-ray can be trained without fake images generated by any of the state-of-the-art face manipulation methods. Extensive experiments show that face X-ray remains effective when applied to forgery generated by unseen face manipulation techniques, while
most existing face forgery detection or deepfake detection algorithms experience a significant performance drop.

\end{abstract}

\section{Introduction}

Recent studies have shown rapid progress in facial manipulation, which enables an attacker to manipulate
the facial area of an image and generate a new image, \eg, 
changing the identities or modifying the face attributes.
With the remarkable success in synthesizing realistic faces, it becomes infeasible even for humans
to distinguish whether an image has been manipulated. 
At the same time, these forged images may be abused for malicious purpose, causing severe trust issues and security concerns in our society. Therefore, it is of paramount importance to develop effective methods for detecting face forgery. 

Our focus in this work is the problem of detecting face forgeries, such as those produced by current state-of-the-art face manipulation algorithms including DeepFakes~\cite{deepfake}, Face2Face~\cite{face2face}, FaceSwap~\cite{faceswap_project}, and NeuralTextures~\cite{neuraltexture}. Face forgery detection is a challenging problem because in real-world scenarios, we often need to detect forgery without knowing the underlying face manipulation methods.
Most existing works~\cite{ding2019swapped,	tariq2018detecting, marra2018detection,rossler2018faceforensics, rossler2019faceforensics++} detect face manipulation in a supervised fashion and their methods are trained for known face manipulation techniques. For such face manipulation, these detection methods work quite well and reach around $98\%$ detection accuracy. However, these detection methods tend to suffer from overfitting and thus their effectiveness is limited to the manipulation methods they are specifically trained for. When applied to forgery generated by unseen face manipulation methods, these detection methods experience a significant performance drop.

Some recent works~\cite{xuan2019generalization,du2019towards} have noticed this problem
and attempted to capture more intrinsic forgery evidence to improve the generalizability.
However, their proposed methods still rely on the generated face forgeries for supervision,
resulting in limited generalization capability.

\begin{figure}
	\centering
	\includegraphics[width=1.0\linewidth]{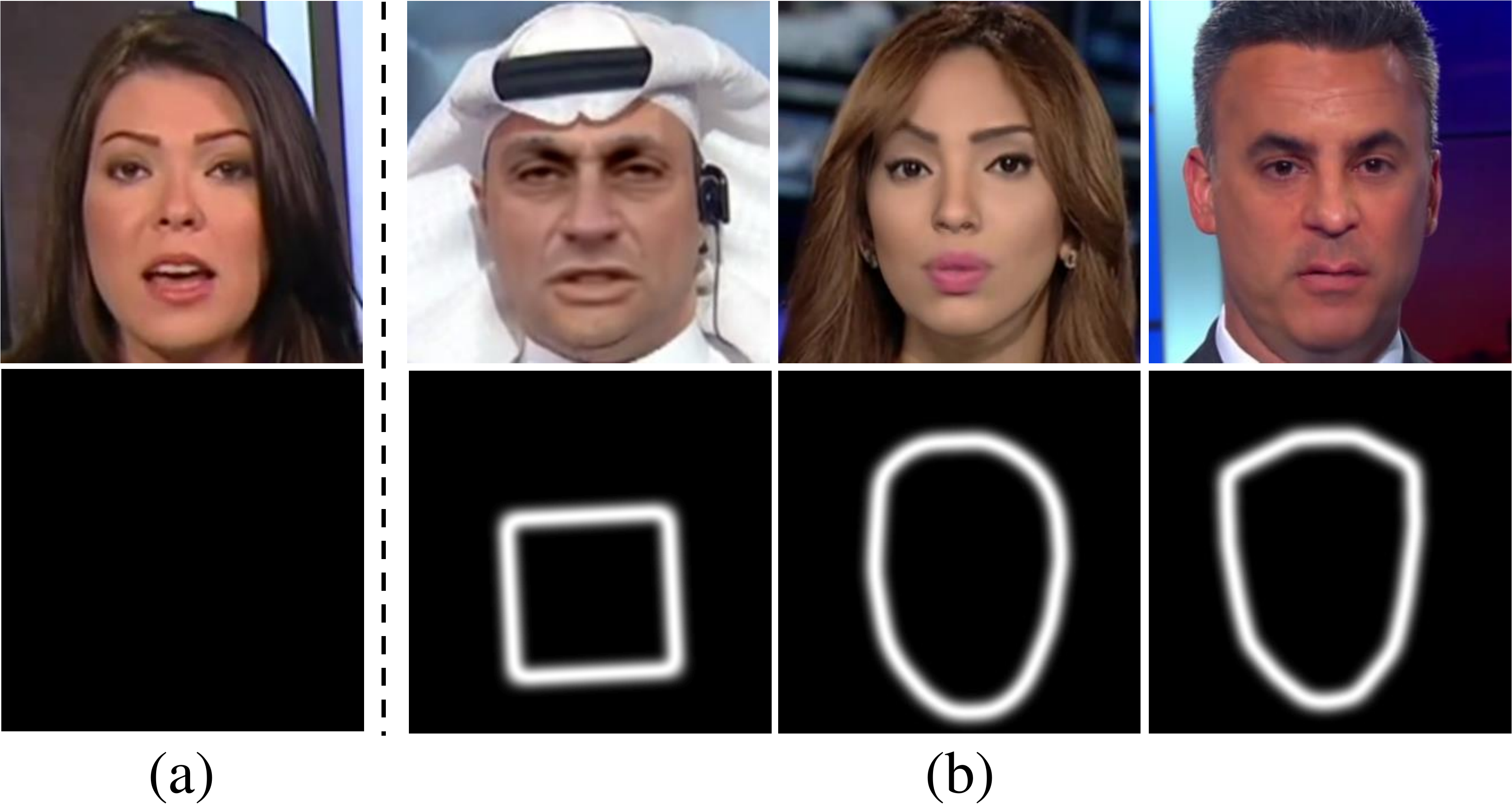}
	\caption{ Face X-ray reveals the blending boundaries in forged face images and returns a blank image for real images. (a) a real image and its face X-ray, (b) fake images and their face X-rays. }
	\label{fig:face-xay}
	\vspace{-0.6cm}
\end{figure}

\begin{figure}
	\centering
	\includegraphics[width=0.95\linewidth]{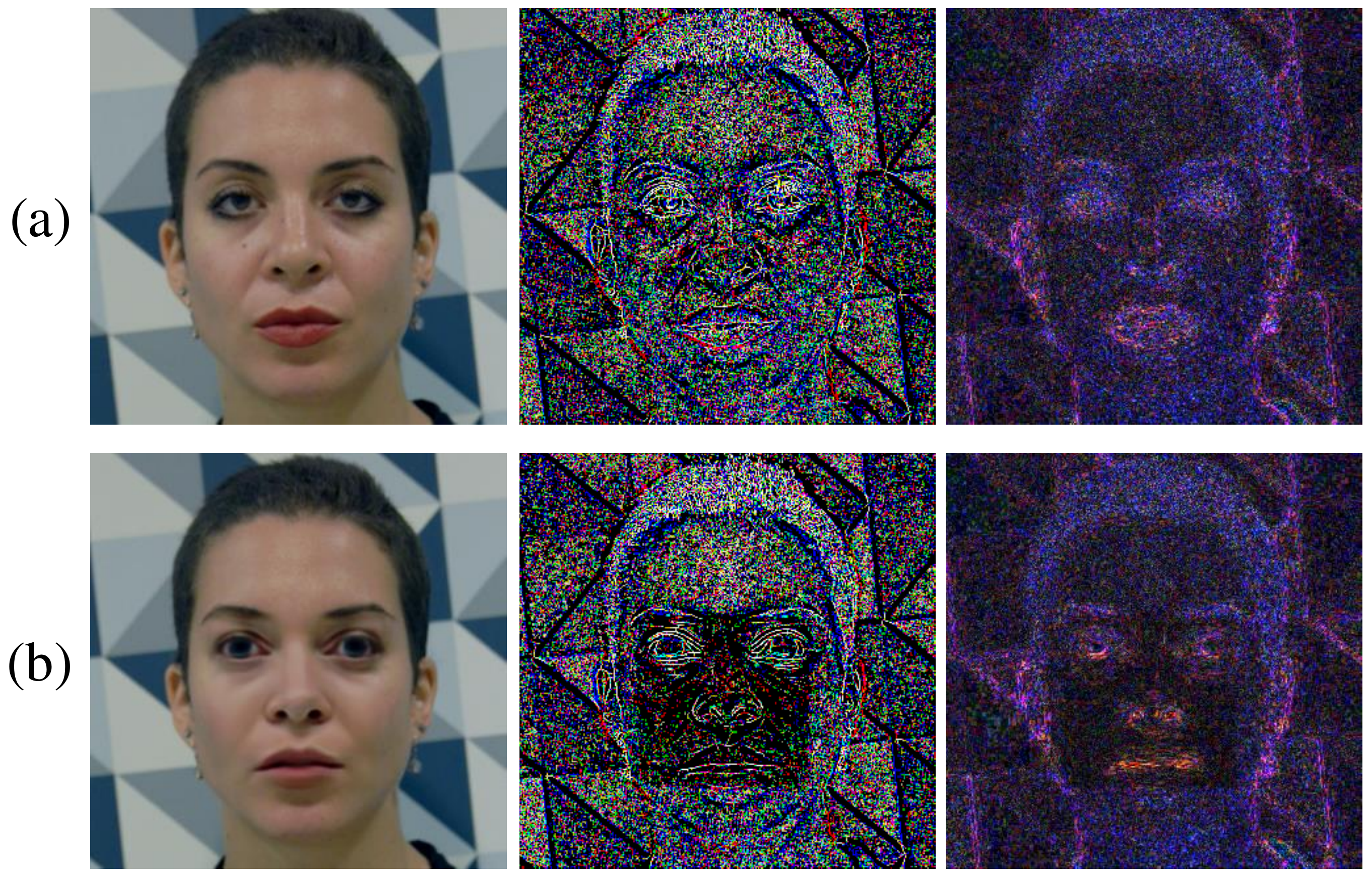}
	\caption{Noise analysis (middle column) and error level analysis (right column) of (a) a real image and (b) a fake image.}
	\label{fig:real_fake_noise_pattern}
	\vspace{-0.6cm}
\end{figure}

In this paper we propose a novel image representation, \emph{face X-ray}, for detecting fake face images or deepfakes. The key observation behind face X-ray is that  most existing face manipulation methods share the common step of blending an altered face into an existing background image, and there exist intrinsic image discrepancies across the blending boundaries. These discrepancies make the boundaries fundamentally detectable. Indeed, due to the acquisition process, each image has its own distinctive marks introduced either from hardware (\eg, sensor, lens) or software components (\eg, compression, synthesis algorithm) and those marks tend to present similarly throughout the image~\cite{schetinger2016digital}. We illustrate noise analysis\footnote{https://29a.ch/photo-forensics/\#noise-analysis} and error level analysis~\cite{krawetz2007picture} as two representative types of distinctive marks in Figure~\ref{fig:real_fake_noise_pattern}. 

Face X-ray capitalizes on the above key observation and provides an effective way for detecting forgery produced by most existing face manipulation algorithms. For an input face image, its face X-ray is a greyscale image that can be reliably computed from the input. This greyscale image not only determines whether a face image is forged or real, but also identifies the location of the blending boundary when it exists, as shown in Figure~\ref{fig:face-xay}.

Face X-ray is a significant step forward in the direction of developing a general face forgery detector
as it only assumes the existence of a blending step and does not rely on any knowledge of the artifacts associated with a specific face manipulation algorithm. This level of generality covers most existing face manipulation algorithms. Moreover, the algorithm for computing face X-ray can be trained with a large number of blended images that are composited from only real ones, without fake images generated by any of the state-of-the-art face manipulation methods.
As a result, face X-ray remains effective when applied to forgery generated by an unseen face manipulation method, while most existing face forgery detection or deepfake detection algorithms experience a significant performance drop.


Our experiments demonstrate that face X-ray significantly improves the generalization ability
through a thorough analysis.
We show that our framework achieves a remarkably high detection accuracy on unseen face forgeries,
as well as the ability to predict face X-rays reliably and faithfully
on recent popular face manipulations.
In comparison with other face forgery detectors,
our framework largely exceeds the competitive state-of-the-arts.

\section{Related Work}

Over the past several years,
forgery creation, of particular interest in faces given its wide applications, has recently gained significant 
attention.
With the complementary property of forgery creation and detection,
face forgery detection also becomes an increasingly emerging research area. In this section, we briefly review prior image forensic methods including face forensics to which our method belongs.


\noindent \textbf{Image forgery classification.}
Image forgery detection is mostly regarded as merely a binary (real or forgery) classification problem.
Early attempts~\cite{pan2012exposing, fridrich2012rich, goljan2015cfa} aim to detect forgeries, 
such as copy-move, removal and splicing that once were the most common manipulations, by 
utilizing intrinsic statistics (\eg, frequency domain characteristics) 
of images.
However, it is difficult to handcraft the most suitable and meaningful features.
With the tremendous success of deep learning,
some works~\cite{cozzolino2017recasting, rahmouni2017distinguishing, bayar2016deep} adopt neural networks to automatically extract discriminative features for forgery detection.

Recent advanced manipulation techniques,
especially about faces, are capable of manipulating the images in a way 
that leaves almost no visual clues and can easily elude
above image tampering detection methods.
This makes face forgery detection increasingly challenging, attracting a large number of
research efforts~\cite{ding2019swapped,	tariq2018detecting, marra2018detection, quan2018distinguishing, mo2018fake, hsu2018learning, afchar2018mesonet,li2018detection,rossler2018faceforensics, rossler2019faceforensics++}.
For instance, a face forensic approach exploiting facial expressions and head movements customized for specific individuals is proposed in~\cite{agarwal2019protecting}.
FakeSpotter
~\cite{wang2019fakespotter} uses layer-wise neuron behavior instead of only the last neuron output to train a binary classifier.
To handle new generated images, an incremental learning strategy is introduced in ~\cite{marra2019incremental}.
Lately, FaceForensics++~\cite{rossler2019faceforensics++} provides 
an extensive evaluation of forgery detectors in various scenarios.


\noindent \textbf{Image forgery localization.}
Besides classification, there are methods focusing on localizing the manipulated region.
Early works~\cite{ryu2013rotation, bianchi2012image, ferrara2012image} reveal the tampered regions 
using manually designed low-level image statistics at a local level.
Subsequently, deep neural network is introduced in image forgery localization,
where most works~\cite{bappy2017exploiting, songsri2019complement, nguyen2019multi, salloum2018image} use multi-task learning to simultaneously 
detect the manipulated images and locate the manipulated region.
Instead of simply using a multi-task learning strategy,
Stehouwer et al.~\cite{stehouwer2019detection} highlight the informative regions through an
attention mechanism where the attention map is guided by 
the groundtruth manipulation mask.
Bappy et al.~\cite{bappy2019hybrid} present a localization architecture that exploits both frequency domain and spatial context.
However, early works are not well suited for detecting advanced manipulations,
while deep learning based methods adopt supervised training,
requiring a huge amount of corresponding groundtruth manipulation masks, which may be inaccessible in practice. 


\begin{figure}[t]
	\centering
	\includegraphics[width=1.0\linewidth]{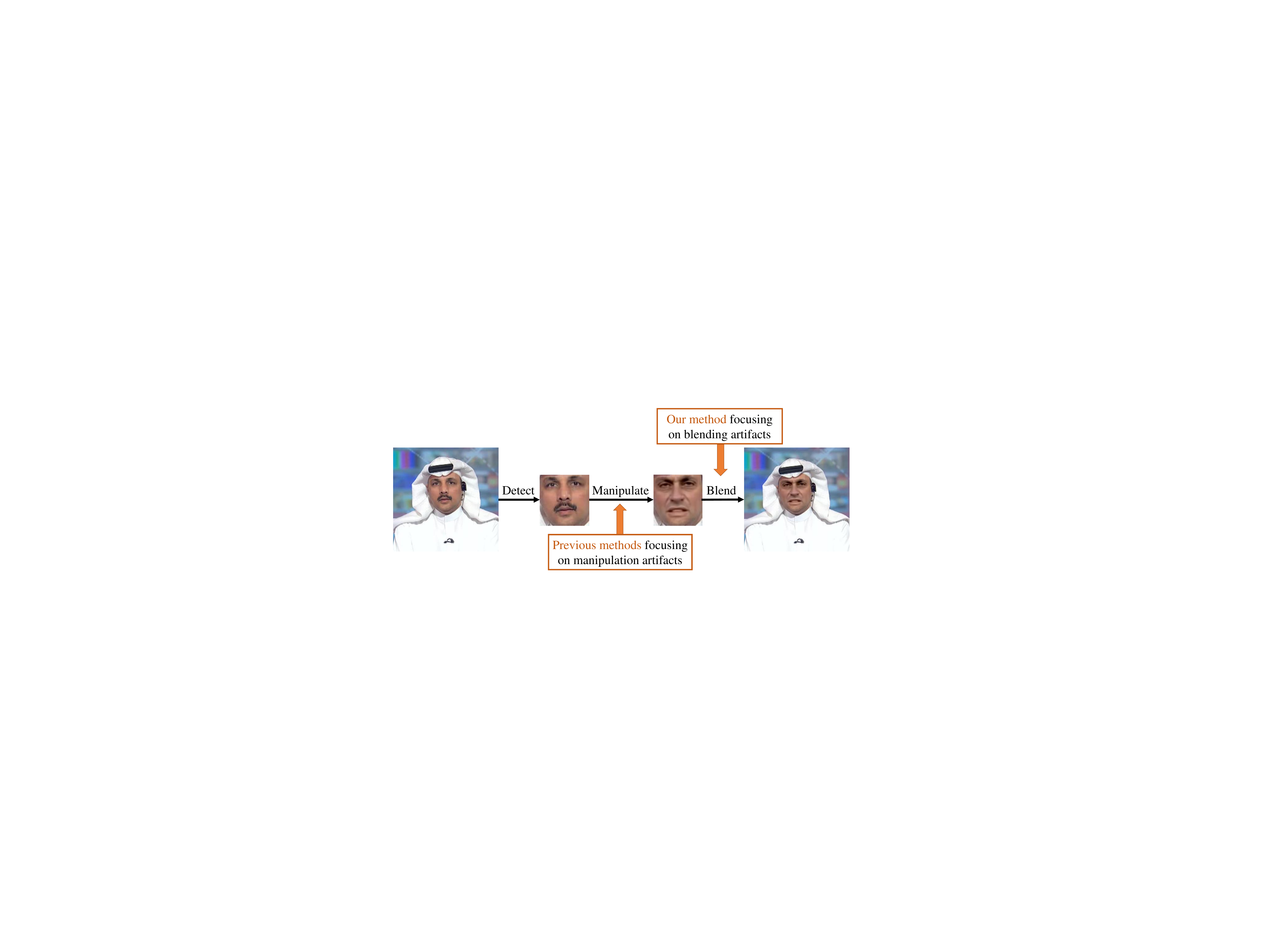}
	\caption{Overview of a typical face manipulation pipeline. Previous works detect artifacts produced by manipulation methods, while our approach focuses on detecting face X-ray.}
	\label{fig:face_manipulated_pipeline}
	\vspace{-0.6cm}
\end{figure}

\noindent \textbf{Generalization ability of Image forgery detection.}
With the evolution of new technologies, it has been noted in recent works~\cite{khodabakhsh2018fake,cozzolino2018forensictransfer,xuan2019generalization, du2019towards, li2018exposing} that the performance of current methods
drop drastically on forgeries of new types.
In particular, Xuan et al.~\cite{xuan2019generalization} use an image preprocessing step to destroy low level unstable artifact, forcing the network to focus on more intrinsic forensic clues.
ForensicTransfer~\cite{cozzolino2018forensictransfer} proposes an autorencoder-based neural network to 
transfer knowledge between different but related manipulations.
LAE~\cite{du2019towards} also uses autorencoder to learn fine-grained representation regularized by forgery mask supervision.
All above methods still need forged images to train a supervised binary classifier, resulting in limited generalization capability.
Another related work is FWA~\cite{li2018exposing}, which targets the artifacts in affine face warping and also can be trained without forged images generated by manipulation methods.
However, FWA focuses on detecting DeepFake generated images such that the detection model is not applicable for other types of manipulations, \eg, Face2Face.

\section{Face X-Ray}


We start by introducing the key observation behind face X-ray. Then we formally define the face X-ray of a given input image.
Finally we provide details on obtaining labeled data (a set of pairs consisting of an image and its corresponding face X-ray) from only real images to train our framework in a supervised manner.

As shown in Figure~\ref{fig:face_manipulated_pipeline}, a typical facial manipulation method consists of three stages: 1) detecting the face area; 2) synthesizing a  desired target face; 3) blending the target face into the original image. 

Existing face forgery detection methods usually focus on the second stage and train a supervised per-frame binary classifier 
based on datasets including both synthesized videos generated from manipulation methods and real ones. Although near perfect detection accuracy is achieved on the test dataset, we observe significant degraded performance when applying the trained model to unseen fake images,
which is empirically verified in Section~\ref{sec:generalization}.

We take a fundamentally different approach. Instead of capturing the synthesized artifacts of specific manipulations in the second stage,
we try to locate the blending boundary that is universally introduced in the third stage.
Our approach is based on a key observation: 
\emph{when an image is formed by blending two images, there exist intrinsic image discrepancies across the blending boundary.}

It is noted in the literature~\cite{farid2009image} that each image has its own distinctive marks or underlying statistics, which mainly come from two aspects:
1) hardware, \eg, color filter array (CFA) interpolation introducing periodic patterns, camera response function that should be similar for each of the color channels, sensor noise including a series of on-chip processings such as quantization and white balancing, introducing a distinct signature;
2) software, \eg, lossy compression schemes that introduce consistent blocking artifacts,
GAN based synthesis algorithms that may leave unique imprints~\cite{marra2019gans,yu2019attributing}.
All above factors contribute to the image formation, leaving specific signatures that tend to be periodic or
homogeneous, which may be disturbed in an altered image. 
As a result, we can detect a forged face image by discovering the blending boundary using the inconsistencies of the underlying image statistics across the boundary.

\begin{figure}[t]
	\centering
	\includegraphics[width=1.0\linewidth]{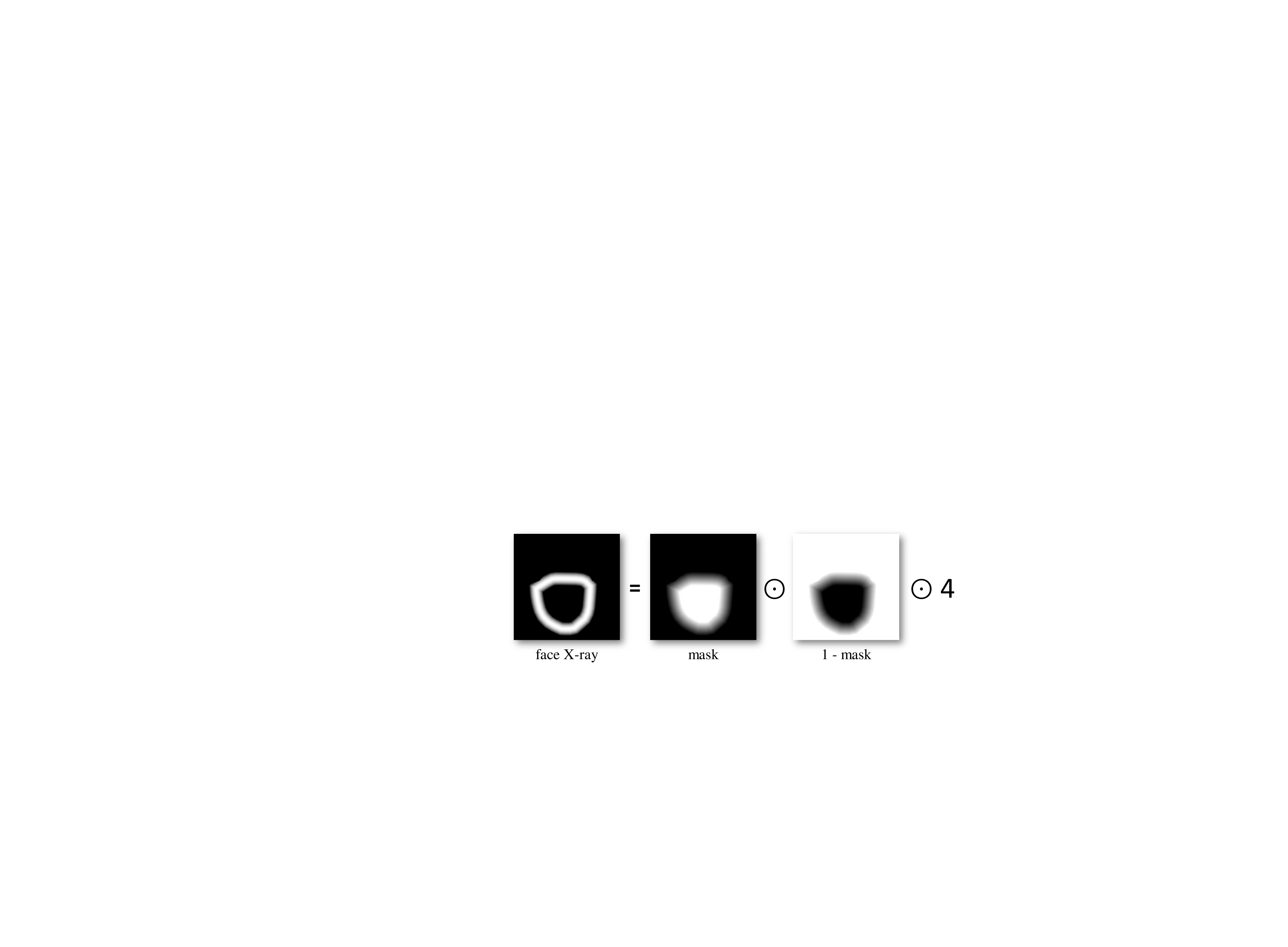}
	\caption{Illustrating the relationship between face X-ray and the mask. $\odot$ represents the element-wise multiplication.}
	\label{fig:boundary}
	\vspace{-0.6cm}
\end{figure}

\subsection{Face X-Ray Definition}
Given an input face image $I$, we wish to decide whether the image is a manipulated image $I_M$ that is obtained by combining two images $I_F$ and $I_B$
\begin{align}
\label{eqn:mask}
I_{M} = M \odot I_F + (1-M) \odot I_B,
\end{align}
where $\odot$ specifies the element-wise multiplication.
$I_F$ is the foreground manipulated face with desired facial attributes, whereas
$I_B$ is the image that provides the background. $M$ is the mask delimiting the manipulated region, with each pixel 
having greyscale value between $0.0$ and $1.0$. When all the entries are restricted to $0$ and $1$, we have a binary mask, such as the mask used in Poisson blending~\cite{perez2003poisson}. Note that color correction techniques (\eg,  color transfer~\cite{reinhard2001color}) are usually applied over the foreground image $I_F$ before blending so that its color matches the background image color.

\begin{figure*}[t]
	\centering
	\includegraphics[width=1.0\linewidth]{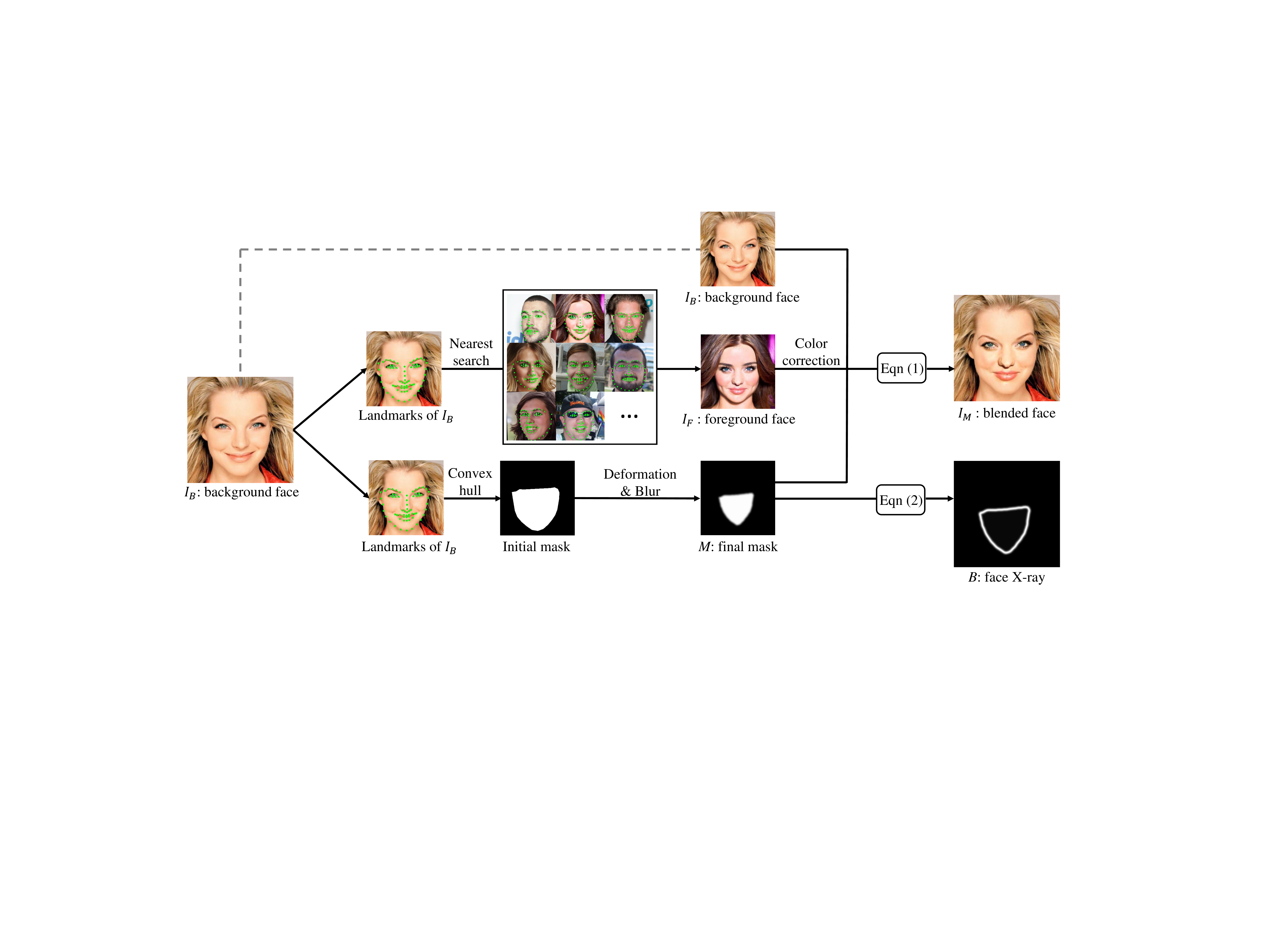}
	\caption{Overview of generating a training sample. Given a real face $I_B$, we seek another real face $I_F$ to represent the manipulated variant of $I_B$ and produce a mask to delimit the manipulated region. Then the blended face and its corresponding face X-ray can be obtained through Equation~(\ref{eqn:mask}) and Equation~(\ref{eqn:boundary}).
		.}
	\label{fig:selfsupervised}
	\vspace{-0.6cm}
\end{figure*}

We would like to define the face X-ray as an image $B$ such that if the input is a manipulated image, $B$ will reveal the blending boundary and if the input is a real image, then $B$ will have zero for all its pixels.

Formally, for an input image $I$, we define its face X-ray as an image $B$ with 
\begin{equation}
\label{eqn:boundary}
B_{i, j} = 4 \cdot M_{i, j} \cdot (1 - M_{i, j}),
\end{equation}
where the subscript $(i,j)$ is the index denoting the pixel location and $M$ is the mask that is determined by the input image $I$. 
If the input image is real, then the mask $M$ is a trivial blank image with either all $0$-pixels or all $1$-pixels. Otherwise, the mask $M$ will be a nontrivial image delimiting the foreground image region. Note that the maximum value of $ M_{i, j} \cdot (1 - M_{i, j})$ is no greater than $0.25$ and in fact only achieves the maximum value of $0.25$ when $M_{i,j}=0.5$. For this reason, the face X-ray pixel $B_{i,j}$ is always valued between $0$ and $1$.
Figure~\ref{fig:boundary} illustrates the relationship between the mask $M$ and the face X-ray $B$ with a toy example. 

In our face X-ray definition, we always assume the mask $M$ is soft and never use a binary mask. A binary mask 
would pose a problem for our face X-ray definition because the corresponding face X-ray would be a blank image with all pixels valued as $0$ even when $M$ is not a trivial mask with all $0$-pixels or all $1$-pixels. This would defeat the purpose of detecting blending boundary in manipulated images. For this reason, we always adopt a $3\times 3$ Gaussian kernel to turn a binary mask 
into a soft mask before using 
Equation~(\ref{eqn:boundary}). 

In essence, the face X-ray aims to discover a soft mask $M$ with which the input image $I$ can be decomposed into the blending of two images from different sources according to Equation~(\ref{eqn:mask}). As mentioned earlier, images from different sources have undeniable differences that, despite their subtlety and invisibility to human eyes, are intrinsic due to the image acquisition process. 
Face X-ray is a computational representation for discovering such differences in an input face image of unknown origin.

\subsection{Training Data Generation from Real Images}
\label{sec:selfsupervised}
Now that we have defined the concept of face X-ray, we shall explain one important thing in the rest of this section: how to get training data using only real images.

As is mentioned before, all real images naturally have their corresponding face X-rays with all $0$-pixels.
Yet those trivial face X-rays are not sufficient to guide the network learning,
training data associated with nontrivial face X-rays is certainly crucial and indispensable.
One intuitive solution is to access the manipulated images and the corresponding masks generated by facial manipulation methods.
Nonetheless, we find that as face X-ray essentially cares about only the blending boundary, it is entirely possible
to create nontrivial face X-rays by blending two real images.

To be specific, we describe the generation of nontrivial face X-rays in three stages.
1) First, given a real image $I_B$, we seek another real image $I_F$ to take the place of the manipulated variant of $I_B$.
We use the face landmarks (extracted by~\cite{chen2014joint}) as the matching criteria to search from a random subset of the rest training videos according to the Euclidean distance between landmarks.
To increase the randomness, we take $100$ nearest neighbors and randomly choose one as the foreground image $I_F$.
2) In the second stage, we generate a mask to delimit the manipulated region.
The initial mask is defined as the convex hull of the face landmarks in $I_B$.
As face manipulation methods are not necessarily and always focused on the same area of the face,
there exist various different shapes of manipulated regions in forged images, \eg, some may be manipulated only around the mouth region.
In order to cover as many shapes of masks as possible,
we first adopt random shape deformation using the 2-D piecewise affine transform estimated from the source $16$ points (selected from a $4\times 4$ grid) to the target $16$ points (deformed from source using random offset),
and then apply Gaussian blur with random kernel size, resulting in the final mask.
3) At last, the blended image is obtained through Equation~(\ref{eqn:mask}), given the foreground image $I_F$, the background image $I_B$ and the mask,
and the blending boundary is attained by Equation~(\ref{eqn:boundary}) using the mask.
Note that we apply the color correction technique (aligning the mean of the RGB channels respectively) to $I_F$, similar to existing facial manipulation methods, so as to match the color of $I_B$.
A brief overview of generating a training sample is illustrated in Figure~\ref{fig:selfsupervised}.
In practice,
we generate 
the labeled data dynamically along with the training process.

\section{Face Forgery Detection Using Face X-Ray} 

As described above,  we are able to produce 
a huge number of training data by exploring only real images.
Let the generated training dataset be $\mathcal{D}=\{I,B,c\}$
where $I$ represents the image, $B$ denotes the corresponding face X-ray
and $c$ is a binary scalar specifying whether the image $I$ is real or blended.
we adopt a convolutional neural network based framework due to the extremely powerful representation learning of deep learning.
The proposed framework outputs the face X-ray given an input image $I$
and then based on the predicted face X-ray, outputs the probabilities of the input image being real or blended.

Formally, let $\hat{B} = NN_b(I)$ be the predicted face X-ray where $NN_b$ is a fully convolutional neural network,
and $\hat{c} = NN_c(\hat{B})$ is the predicted probability with $NN_c$ composed of a global average pooling layer, a fully connected layer, and a softmax activation layer in a sequential manner.
During training, 
we adopt the widely used loss functions for the two predictions.
For face X-ray, we use the cross entropy loss to measure the accuracy of the prediction,
\begin{equation} 
\scriptsize
L_{b} = - \sum_{\{I,B\}\in \mathcal{D}}\frac 1 N \sum_{i,j} (B_{i,j} \text{log} \hat{B}_{i,j} + (1-B_{i,j}) \text{log} (1-\hat{B}_{i,j})),
\end{equation} 
where $N$ is the total number of pixels in the feature map $\hat{B}$.
For classification,
the loss is
\begin{align}
L_{c} = -\sum_{\{I,c\}\in \mathcal{D}}(c \text{log}(\hat{c}) + (1-c)\text{log}(1-\hat{c})).
\end{align}
Therefore,
the overall loss function is $L = \lambda L_b + L_c$, where $\lambda$ is the loss weight balancing $L_b$ and $L_c$. In the experiments, we set $\lambda = 100$ to force the network focusing more on learning the face X-ray prediction.
We train our framework in an end-to-end manner using the back propagation.
More implementation details can be found in Section~\ref{sec:experiments}.


\section{Experiments}
\label{sec:experiments}

In this section,
we first introduce the overall experiment setups
and then present extensive experimental results to demonstrate the superiority of our approach.

\noindent \textbf{Training datasets.}
In our experiments, we adopt recent released benchmark dataset FaceForensics++~\cite{rossler2019faceforensics++} (FF++) for training.
It is a large scale video dataset consisting of
$1000$ original videos that have been manipulated with four state-of-the-art face manipulation methods:
DeepFakes (DF)~\cite{deepfake}, Face2Face (F2F)~\cite{face2face}, FaceSwap (FS)~\cite{faceswap_project},
and NeuralTextures (NT)~\cite{neuraltexture}.
Another training dataset is the set of blended images that we constructed from real images.
We denote such dataset with BI, meaning the blended data samples composited using real images in FF++.

\noindent \textbf{Test datasets.}
To evaluate the generalization ability of the proposed model using face X-ray,
we use the following datasets:
1) FaceForensics++~\cite{rossler2019faceforensics++} (FF++) that contains four types of facial manipulations as described above;
2) DeepfakeDetection\footnote{https://ai.googleblog.com/2019/09/contributing-data-to-deepfake-detection.html} (DFD) including thousands of visual deepfake videos released by Google in order to support developing deepfake detection methods;
3) Deepfake Detection Challenge\footnote{https://deepfakedetectionchallenge.ai/dataset}
(DFDC) released an initial dataset of deepfakes accompanied with labels describing 
whether they are generated using facial manipulation methods;
4) Celeb-DF~\cite{li2019celeb}, a new DeepFake dataset including 408 real videos and 795 synthesized video with reduced visual artifacts.


\noindent \textbf{Implementation detail.}
For the fully convolutional neural network $NN_b$ in our framework, we adopt the recent advanced neural network architecture, \ie, HRNet~\cite{sun2019deep,sun2019high}, and then concatenate representations from all four different resolutions to the same size $64\times 64$, followed by a $1\times 1$ convolutional layer with one output channel, a bilinear upsampling layer with $256\times 256$ output size and a sigmoid function.
In the training process, the batch size is set to $32$ and the total number of iterations is set to $200,000$.
To ease the training process of our framework,
we warm start the remaining layers with fixed ImageNet pre-trained HRNet for the first $50,000$ iterations
and then finetune all layers together for the rest $150,000$ iterations.
The learning rate is set as $0.0002$ using Adam~\cite{kingma2014adam} optimizer at first and then is linearly decayed to $0$ for the last $50,000$ iterations.

\begin{table}[t]
	\setlength\tabcolsep{4.5pt} 
		\footnotesize
	\begin{center}
		\begin{tabular}{c|cc|cccc|c}
			\hline
			\multirow{2}{*}{Model} &	\multicolumn{2}{c|}{	Training set} & \multicolumn{5}{c}{Test set AUC}
			 \\
			\cline{2-8}
			&\textcolor{brown} {DF}& BI	& \textcolor{brown} {DF}& F2F & FS & NT & FF++
			\\
			\hline
			Xception~\cite{rossler2019faceforensics++} & \checkmark& -- & \textcolor{brown} {99.38} & 75.05 & 49.13 & 80.39  & 76.34\\
			HRNet&	\checkmark & -- & \textcolor{brown} {99.26} & 68.25 & 39.15 & 71.39 & 69.51 \\ 
			\hline  
			\multirow{2}{*}{	Face X-ray} & \checkmark & --& 
			\textcolor{brown} {99.17} & 94.14 & 75.34 & 93.85 &90.62 \\ 
			& \checkmark &  \checkmark & \textcolor{brown} {99.12} & \textbf{97.64}  & \textbf{98.00} & \textbf{97.77}  & \textbf{97.97}\\ 
			\hline
			\hline
			&\textcolor{brown} {F2F}& BI	& DF& \textcolor{brown} {F2F} & FS & NT & FF++
			\\
			\hline
			Xception~\cite{rossler2019faceforensics++} & \checkmark & -- & 87.56 & \textcolor{brown} {99.53} &  65.23 & 65.90 & 79.55 \\
			HRNet&	\checkmark & -- &  83.64 & \textcolor{brown} {99.50} &  56.60 & 61.26 & 74.71 \\ 
			\hline  
			\multirow{2}{*}{	Face X-ray} & \checkmark & -- 
			& 98.52 & \textcolor{brown} {99.06} & 72.69  &  91.49 &93.41\\ 
			& \checkmark &  \checkmark &  \textbf{99.03}& \textcolor{brown} {99.31} & \textbf{98.64} & \textbf{98.14} & \textbf{98.78} \\ 
			\hline
			\hline
			&\textcolor{brown} {FS}& BI	& DF& F2F & \textcolor{brown} {FS} & NT & FF++
			\\
			\hline
			Xception~\cite{rossler2019faceforensics++} & \checkmark& --  & 70.12 & 61.70 &\textcolor{brown} {99.36} & 68.71  & 74.91 \\ 
			HRNet&	\checkmark & -- & 63.59 & 64.12 &\textcolor{brown} {99.24} & 68.89&73.96\\
			\hline  
			\multirow{2}{*}{	Face X-ray} & \checkmark & --
			& 93.77 & 92.29 & \textcolor{brown} {99.20} & 86.63 &93.13\\ 
			& \checkmark &  \checkmark & \textbf{99.10} & \textbf{98.16} & \textcolor{brown} {99.09} & \textbf{96.66} & \textbf{98.25} \\ 
			\hline
			\hline
			&\textcolor{brown} {NT}& BI	& DF& F2F & FS & \textcolor{brown} {NT} & FF++
			\\
			\hline
			Xception~\cite{rossler2019faceforensics++} & \checkmark& -- & 93.09 & 84.82 & 47.98 & \textcolor{brown} {99.50} &  83.42\\
			HRNet&	\checkmark & -- & 94.05 & 87.26 & 64.10 & \textcolor{brown} {98.61} & 86.01 \\ 
			\hline  
			\multirow{2}{*}{	Face X-ray} & \checkmark & --&  99.14 & 98.43 & 70.56 & \textcolor{brown} {98.93} & 91.76\\ 
			& \checkmark &  \checkmark & \textbf{99.27} & \textbf{98.43} & \textbf{97.85} & \textcolor{brown} {99.27} & \textbf{98.71}\\ 
			\hline
			\hline
			& FF++ & BI & DF & F2F & FS & NT & FF++ \\
			\hline
			
			Xception~\cite{rossler2019faceforensics++}& -- & \checkmark & 98.95 & 97.86& 89.29 & 97.29 & 95.85\\
			HRNet & --& \checkmark& 99.11 & 97.42& 83.15& \textbf{98.17} & 94.46  \\
	
			\hline
			Face X-ray & -- & \checkmark & \textbf{99.17}& \textbf{98.57}& \textbf{98.21}& 98.13 & \textbf{98.52}\\
			\hline

		\end{tabular}
	\end{center}
\caption{Generalization ability evaluation. Using only classifier suffers performance drop on unseen facial manipulations. Our approach improves the generalization ability by detecting face X-ray and further gets significant improvement using the constructed BI dataset.
	It is worth noting that our framework only using BI dataset still obtains promising results.} 
\label{tab:boundarydetection}
\vspace{-0.7cm}
\end{table}

\subsection{Generalization Ability Evaluation}
\label{sec:generalization}

We first verify that supervised binary classifiers experience a significant performance drop over unseen fake images.
To show this,
we adopt the state-of-the-art detector, \ie Xception~\cite{rossler2019faceforensics++}.
Table~\ref{tab:boundarydetection} summarizes the results in terms of AUC (area under the Receiver Operating Characteristic curve) with respect to each type of manipulated videos.
We observe that excellent performance (above $99\%$) is obtained on the known specific manipulation,
while the performance drops drastically for unseen manipulations.
The reason may be that the model quickly overfits
to the manipulation-specific artifacts, achieving high performance for the given data but suffering from poor generalization ability.

Our approach tackles the forgery detection by using a more general evidence: face X-ray.
We show that the improved generalization ability comes from two factors:
1) we propose detecting the face X-ray instead of paying attention to the manipulation-specific artifacts;
2) we construct a large number of training sample automatically and effortlessly composited from real images so that the model
is adapted to focus more on the face X-rays.
Finally, we show that our method, even only using the constructed BI data, is capable of achieving a high detection accuracy.

\noindent \textbf{The effect of detection using face X-ray.}
We first evaluate our model detecting face X-rays using the same training set and training strategy as Xception~\cite{rossler2019faceforensics++}.
In order to obtain accurate Face X-rays for the manipulated images, 
we again adopt the generation process in Section~\ref{sec:selfsupervised} by considering the real image as background and the fake image as foreground, given a pair of a real image and a fake one.
For fair comparison, we also show the results of 
the binary classifier using the same network architecture with ours, which is denoted as HRNet in the table.
The comparison results are shown in Table~\ref{tab:boundarydetection}.
It can be clearly seen that our approach gets significant improvement on the unseen facial manipulations,
verifying our hypothesis that explicitly detecting face X-ray is more generalizable.

\noindent \textbf{The effect of additional blended images.}
Further, we train our framework with additional blended images that capture various types of face X-rays.
The results are given in Table~\ref{tab:boundarydetection},
showing that large improvement has been obtained again.
We think that there are two advantages.
One is the benefit of extra training data as
it is known that increasing the amount of training data always leads to better model and thus improved performance.
Another important thing is that the region inside the boundary of blended images is actually real instead of synthesized, making the model less over-fitting to manipulation-specific artifacts.

\noindent \textbf{Results of only using the blended images.}
Finally we present the results of our framework using only the blended images in 
Table~\ref{tab:boundarydetection}. 
The performance in terms of AUC on the four representative facial manipulations DF, F2F, FS, NT
is $99.17\%, 98.57\%, 98.21\%, 98.13\%$ respectively.
This shows that our model, even without fake images generated by any of the state-of-the-art face manipulation methods, still achieves a competitive high detection accuracy.
We also show the results of classifer which is trained over BI by considering BI as fake images.
The performance is much better than the classifiers trained with forged images generated by manipulation methods.
This is probably because BI forced the classifier to learn the face X-rays, leading to better generalization.
Nevertheless, our approach using face X-ray still gets overall better results and thus again validates the conclusion that using face X-ray is more generalizable.

\begin{figure*}[t]
	\centering
	\includegraphics[width=1.0\linewidth]{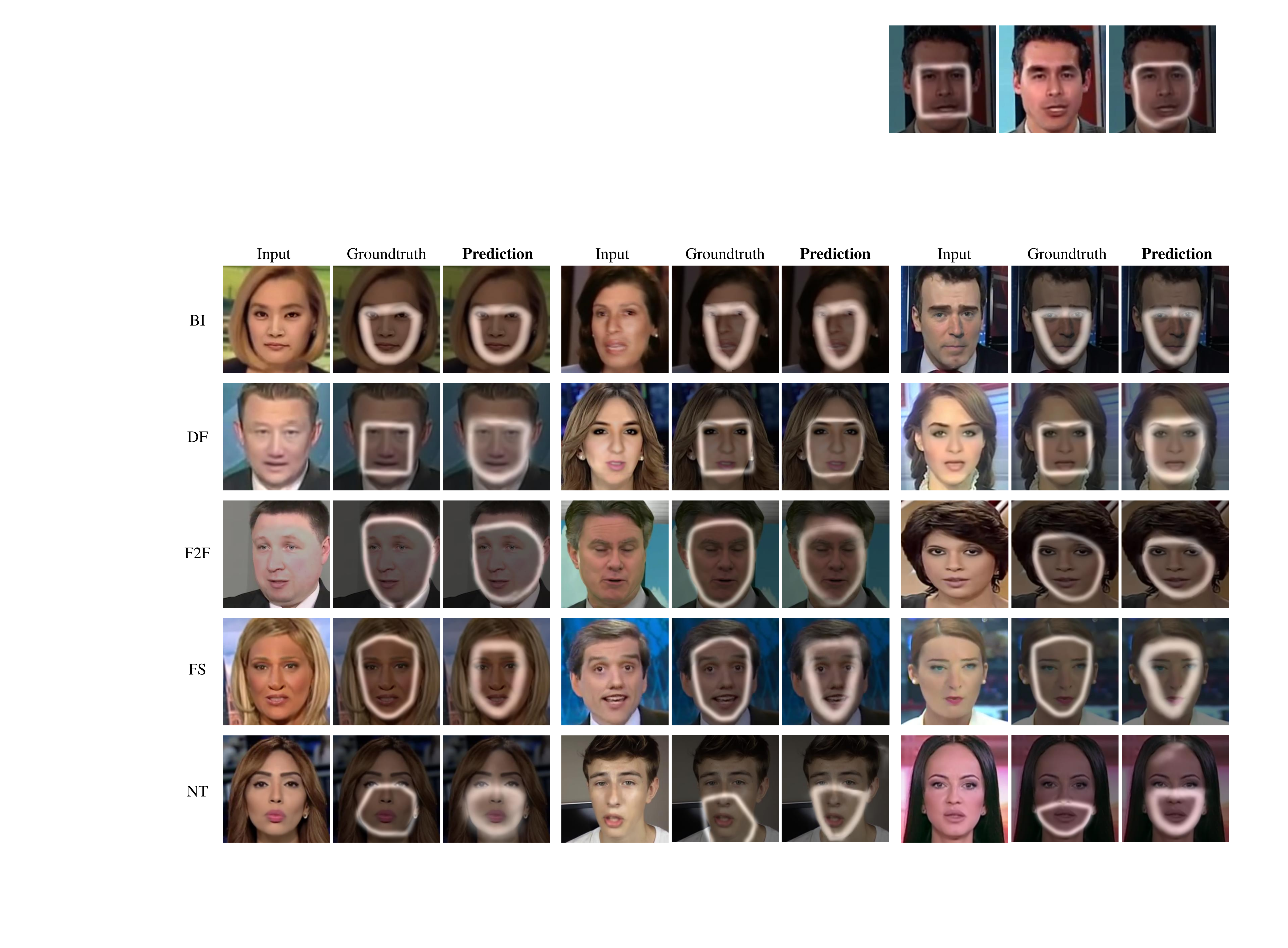}
	\caption{Visual results on various facial manipulation methods including our generated blended images. For the facial manipulations,
		the groundtruth is obtained by computing the absolute element-wise difference between the manipulated image and the corresponding real image and then converting to grayscale followed by normalization.
		It can be clearly seen from the figure that
		the predicted face X-ray well captures the shape of the corresponding groundtruth.
	More visual results can be found in the supplementary.}
	\label{fig:our_boundary_predict_results}
\end{figure*}

\begin{table*}[t]
	
	\begin{center}
		\begin{tabular}{c|c|ccc|ccc|ccc}
			\hline
			\multirow{2}{*}{Model} &
			\multirow{2}{*}{Training dataset} & \multicolumn{9}{c}{Test dataset} \\
			\cline{3-11}
			&	&	 \multicolumn{3}{c|}{DFD} & \multicolumn{3}{c|}{DFDC} & \multicolumn{3}{c}{Celeb-DF} 
			\\
			\hline
			\multicolumn{2}{c|}{}  & AUC & AP & EER &AUC & AP & EER &AUC & AP & EER \\
			\hline
			Xception~\cite{rossler2019faceforensics++} & FF++& 87.86 & 78.82 & 21.49 & 48.98 & 50.83 & 50.45  & 36.19 & 50.07 & 59.64 \\
			\hline 
			Face X-ray &BI & 93.47 & 87.89 & 12.72 &  71.15 & \textbf{73.52} & 32.62 & 74.76 & 68.99 & 31.16\\
			Face X-ray &FF++ and BI  & \textbf{95.40} & \textbf{93.34} & \textbf{8.37} & \textbf{80.92} & 72.65 & \textbf{27.54} & \textbf{80.58} & \textbf{73.33} & \textbf{26.70}\\
			\hline
		\end{tabular}
	\end{center}
\caption{Benchmark results in terms of AUC, AP and EER for our framework and the state-of-the-art detector Xception~\cite{rossler2019faceforensics++} on unseen datasets.
Our framework, trained using the blended images, 
already performs better than the baseline.
The performance is further greatly improved in most cases if we exploit additional fake images even not from the same distribution as the test set.} 
	\label{tab:benchmark}
	\vspace{-0.6cm}
\end{table*}

\subsection{Benchmark Results on Unseen Datasets}


In order to facilitate the industry as well as the academia to develop advanced face forgery detectors,
a growing number of datasets containing a large amount of high-quality deepfake videos 
have been released recently.
Here we show the benchmark results of our framework
on the detection of those unseen popular datasets.

\noindent \textbf{Results in terms of forgery classification.}
We first show the forgery classification results in terms of AUC, AP (Average Precision)  and EER (Equal Error Rate).
The results of our framework on recent released large scale datasets, \ie DFD, DFDC and Celeb-DF, are shown in Table~\ref{tab:benchmark}.
We also show the results of the state-of-the-art detector Xception~\cite{rossler2019faceforensics++} as a baseline.
We can see that 
our framework, without using 
any images generated from facial manipulation methods, 
already performs better than the baseline.
Moreover, if we exploit additional fake images even not from the same distribution as the test set, the performance is further greatly improved in most cases.

\noindent \textbf{Results in terms of face X-ray prediction.}
Our framework makes predictions about the forgery based on the existence of nontrivial face X-rays.
We show that our method can reliably predict the face X-rays for unseen facial manipulations
and thus provide explainable decisions for the inference.
The visual examples on various types of fake images including the blended images generated in the proposed process are shown in Figure~\ref{fig:our_boundary_predict_results}.
For the facial manipulations,
the groundtruth is obtained by computing the absolute element-wise difference between the manipulated image and the corresponding real image and then converting to grayscale followed by normalization.
It can be clearly seen from the figure that
the predicted face X-ray well captures the shape of the corresponding groundtruth.



\subsection{Comparison with Recent Works}

Some recent related works~\cite{li2018exposing,du2019towards,cozzolino2018forensictransfer,nguyen2019multi} have also 
noticed the generalization issue and tried to solve the problem to a certain degree.
FWA~\cite{li2018exposing} also creates negative samples from real images.
Yet its goal is to characterize face warping artifacts only widely existed in DeepFake generated videos.
The comparison is given in Table~\ref{tab:comparisonFWA}.
Three other related works are LAE~\cite{du2019towards}, FT~\cite{cozzolino2018forensictransfer},
both attempting to learn intrinsic representation instead of capturing artifacts in the training set,
and MTDS~\cite{nguyen2019multi} learning detection and localization simultaneously.
We present the comparison, which is evaluated over a new type of manipulated data when the model is trained on another type, in Table~\ref{tab:comparison}.
Note that we directly cite the numbers from their original papers for fair comparison.
From the two tables, we can see that our framework largely exceeds recent state-of-the-arts.

\begin{table}[t]
	\setlength\tabcolsep{12pt} 
	\begin{center}
		\begin{tabular}{c|cc}
			\hline
			\multirow{2}{*}{Model} &  \multicolumn{2}{c}{AUC} \\
			
			\cline{2-3}
			&  FF++/DF & Celeb-DF \\
			\hline
			FWA~\cite{li2018exposing}  &79.20 & 53.80 \\
			\hline
			Face X-ray & \textbf{99.17} & \textbf{74.76} \\
			\hline
		\end{tabular}
	\end{center}
	\caption{AUC comparison with FWA.} 
	\label{tab:comparisonFWA}
	\vspace{-0.6cm}
\end{table}

\begin{table}[t]
	
	\begin{center}
		\begin{tabular}{c|cc|cc}
			\hline
			\multirow{2}{*}{Model} &	\multicolumn{2}{c|}{	Training set} & \multicolumn{2}{c}{Detection accuracy} \\
			
			\cline{2-5}
			&		F2F & FS	& F2F & FS \\
			\hline
			LAE~\cite{du2019towards} &	 \checkmark & -- & 90.93 & 63.15 \\
			\hline  
			FT-res~\cite{cozzolino2018forensictransfer} & \checkmark & 4 images & 94.47 & 72.57 \\
			\hline
			MTDS~\cite{nguyen2019multi} & \checkmark & -- & 92.77 & 54.07 \\
			\hline
			Face X-ray	& \checkmark & -- & \textbf{97.73} & \textbf{85.69}  \\
			\hline
		\end{tabular}
	\end{center}
	\caption{Detection accuracy comparison with recent methods. Note that here we use the HQ version (a light compression) of FF++ dataset for fair comparison.} 
	\label{tab:comparison}
	\vspace{-0.7cm}
\end{table}

\subsection{Analysis of the proposed framework}

\noindent \textbf{The effect of data augmentation.}
The overall goal of data augmentation in the training data generation is to offer a large amount of different types of blended images to give the model the ability to detect various manipulated images.
Here, we study two important augmentation strategies: a) mask deformation which intends to bring larger variety to the shape of face X-ray; 
b) color correction in order to produce a more realistic blended image.
We think that both strategies are crucial for generating diverse and high-quality data samples that are definitely helpful for network training.
To show this, 
we present the comparison on FF++ and DFD in Table~\ref{tab:fake_data_generation_ablation_study}.
It can be seen that both strategies are important and without either one
will degrade the performance.

\noindent \textbf{The effect of the loss weight $\lambda$.}
In the loss function, we impose a loss weight $\lambda$ to balance the classification loss and the face X-ray prediction loss and set it as $100$ in the experiments. Here we show the results when using different values of $\lambda$ in Table~\ref{tab:comparisonlambda}. We have two observations:
(1) the results when $\lambda$ is set as $10$, $100$ or $1000$ are comparable (with AUC being $98.4\%$, $98.5\%$, $98.6\%$ respectively);
(2) the performance when $\lambda$ is set as non-zero is better than the performance when $\lambda$ is zero (with AUC being $94.5\%$), suggesting that the face X-ray prediction loss is helpful.

\noindent \textbf{Generalization to other types of blending.}
We adopt alpha blending in the training data generation.
Though we have demonstrated the performance 
on unseen manipulations that might not use alpha blending,
we here precisely present a study on the results of our approach with respect to Poisson blending,
another widely used blending technique in existing face manipulation methods,
and deep blending (GP-GAN~\cite{wu2019gp}) which utilizes neural network rather than Equation~(\ref{eqn:mask}).
We construct the test data by using different types of blending and evaluate the model when the training data is constructed with alpha blending.
The results are given in Table~\ref{tab:blendingtype}.
We can see that our framework still gets satisfactory results on unseen blending types though with visible performance drops on Possion blending.

\section{Limitations}

While we have demonstrated satisfactory performance on general detection in the experiments,
we are aware that there exist some limitations of our framework.

First, we realize that detecting face X-ray may fail in two aspects.
1) Our method relies on the existence of a blending step.
Therefore, when an image is entirely synthetic, it is possible that our method may not work correctly.
However the fake news
such as videos of someone saying and doing things they didn't, usually require blending as a post-processing step.
This is because so far without blending, it is unlikely to completely generate a realistic image with desired target background. 
We indeed provide a promising way to detect those numerous blended forgeries.
2) We notice that 
one can develop adversarial samples to against our detector.
This is inevitable since it is an arms race between image forgery creation and detection,
which would inspire both fields to develop new and exciting techniques.

Besides, similar to all the other forgery detectors,
our method also suffers from performance drop when encounter low resolution images.
This is because 
classifying low resolution images is more challenging as the forgery evidence is less significant.
We test our framework on the HQ version (a light compression) and the LQ version (a heavy compression) of FF++ dataset and the overall AUC 
are $87.35\%$ and $61.6\%$ respectively.
This is expected as the heavier the compression, the less significant the forgery evidence and
thus the lower the performance.

\begin{table}[t]
	\setlength\tabcolsep{12pt} 
	\begin{center}
		\begin{tabular}{c|cc}
			\hline
			\multirow{2}{*}{} & \multicolumn{2}{c}{AUC} \\
			\cline{2-3}
			& FF++ & DFD \\
			\hline
			w/o mask deformation & 93.92 & 85.89\\
			\hline
			w/o color correction & 96.21 & 89.91 \\
			\hline
			Face X-ray & \textbf{98.52} & \textbf{93.47} \\
			\hline
		\end{tabular}
	\end{center}
	\caption{Ablation study for the effect of (a) mask deformation and (b) color correction in the training data generation pipeline.} 
	\label{tab:fake_data_generation_ablation_study}
	\vspace{-0.2cm}
\end{table}

\begin{table}[t]
	\setlength\tabcolsep{6pt} 
	\begin{center}
		\begin{tabular}{c|ccccc}
			\hline
			$\lambda$  &0 & 1 & 10 & 100 & 1000 \\
			\hline
			AUC & 94.5\% & 97.9\% & 98.4\% & 98.5\% & 98.6\% \\
			\hline
		\end{tabular}
	\end{center}
	\caption{The effect of the loss weight $\lambda$ in the loss function.} 
	\label{tab:comparisonlambda}
	\vspace{-0.3cm}
\end{table}

\begin{table}[t]
	\setlength\tabcolsep{10pt} 
	\begin{center}
		\begin{tabular}{l|ccc}
			\hline
			Blending type &
			AUC & AP & EER\\
			\hline
			Alpha blending & 99.46 & 98.50 & 1.50\\
			\hline
			Possion blending & 94.62 & 88.85 & 11.41 \\
			Deep blending~\cite{wu2019gp} & 99.90 & 98.77 & 1.36\\
			\hline
		\end{tabular}
	\end{center}
	\caption{Results over test data using Possion blending and deep blending when the training data is constructed with alpha blending.
	} 
	\label{tab:blendingtype}
	\vspace{-0.6cm}
\end{table}

\section{Conclusion}

In this work, we propose a novel face forgery evidence, face X-ray, based on the observation that
most existing face manipulation methods share a common blending step
and there exist intrinsic image discrepancies across the blending boundary, which is neglected in advanced face manipulation detectors.
We develop a more general face forgery detector using face X-ray and
the detector can be trained without fake images generated by any
of the state-of-the-art face manipulation methods.
Extensive experiments have been performed to demonstrate the generalization ability of face X-ray,
showing that our framework
is capable of accurately distinguishing unseen forged images and reliably predicting the blending regions.

{\small
\bibliographystyle{ieee_fullname}
\bibliography{egbib}
}

\end{document}